\documentclass[a4paper,twoside]{article}

\usepackage{epsfig}
\usepackage{subcaption}
\usepackage{calc}
\usepackage{amssymb}
\usepackage{amstext}
\usepackage{amsmath}
\usepackage{amsthm}
\usepackage{multicol}
\usepackage{pslatex}
\usepackage{apalike}
\usepackage{algorithm2e}
\usepackage{url}
\usepackage[bottom]{footmisc}
\usepackage{SCITEPRESS}     % Please add other packages that you may need BEFORE the SCITEPRESS.sty package.

\usepackage{tcolorbox}
\usepackage[table]{xcolor}

\begin{document}

\title{SAC: A Framework for Measuring and Inducing Personality Traits in LLMs with Dynamic Intensity Control}

\author{\authorname{Adithya Chittem\sup{1}, Aishna Shrivastava\sup{1}, Sai Tarun Pendela\sup{1}, Jagat Sesh Challa\sup{1}, Dhruv Kumar\sup{1}}
\affiliation{\sup{1} BITS Pilani, India}
\email{\{f20220012, f20220006, f20220419, jagatsesh, dhruv.kumar\}@pilani.bits-pilani.ac.in}
}

% \keywords{The paper must have at least one keyword. The text must be set to 9-point font size and without the use of bold or italic font style. For more than one keyword, please use a comma as a separator. Keywords must be titlecased.}
\keywords{Large Language Models,
Personality Modelling,
16 Personality Factor (16PF) Model,
Trait Intensity Control,
Machine Personality Inventory (MPI),
Human-Machine Interaction}

% # TODO:
% elaborate on related work
% take interviews 
% human baseline section

% \abstract{The abstract should summarize the contents of the paper and should contain at least 70 and at most 200 words. The text must be set to 9-point font size.}

\abstract{Large language models (LLMs) have rapidly gained traction across fields, with growing expectations for them to exhibit human-like personalities. To meet this expectation, various studies have explored personality modelling through psychometric frameworks. However, most existing approaches rely on the Big Five (OCEAN) model, which offers only broad personality dimensions and lacks mechanisms for fine-grained control of trait intensity.
We address this gap by extending the Machine Personality Inventory (MPI), which originally used the Big Five model, to incorporate the 16 Personality Factor (16PF) model, allowing expressive control over sixteen distinct traits. We also developed a structured framework known as Specific Attribute Control (SAC) for evaluating and dynamically inducing trait intensity in LLMs.
Our method introduces adjective-based semantic anchoring to guide trait intensity expression and leverages behavioural questions across five intensity factors: \textit{Frequency}, \textit{Depth}, \textit{Threshold}, \textit{Effort}, and \textit{Willingness}. 
We tested this methodology on four state-of-the-art LLMs and benchmarked it against human responses, finding that modelling intensity as a continuous spectrum yields more consistent and controllable personality expression than binary trait toggling.
We also find that altering a target trait’s intensity systematically shifts related traits in psychologically coherent directions, indicating that LLMs internalize multi-dimensional personality structures rather than treating traits independently. These findings open new directions for developing LLMs with adaptive, transparent, and context-sensitive personality control.}

% used three open source models
% experimental results demonstrate evaluation metrics are suitable for evaluating LLMs’ personality in terms of stability and tendency.
\onecolumn \maketitle \normalsize \setcounter{footnote}{0} \vfill
\section{\uppercase{Introduction}}
\label{sec:introduction}

% Your paper will be part of the conference proceedings therefore we ask that authors follow the guidelines explained in this example in order to achieve the highest quality possible \cite{Smith98}.

% Be advised that papers in a technically unsuitable form will be returned for retyping. After returned the manuscript must be appropriately modified.
Personality can be broadly defined as the characteristic patterns of thought, emotion, and behavior that influences how individuals perceive the world, respond to others, and navigate complex social dynamics \cite{mischel2004integrative}. In human interaction, personality shapes expectations, trust, empathy, and social relationships \cite{back2023personalitysocialrelationships}. As artificial agents begin to take on roles that require emotional intelligence and social fluency, personality becomes a central axis along which user perception and comfort are determined.

LLMs are increasingly being utilized in domains where effective human interaction is essential, including education \cite{alhafni2024llmsineducation}, healthcare \cite{cascella2023chatgpthealthcare}, and creative applications \cite{kim2025evaluatingcreativity}. In such contexts, users not only seek accurate responses but also value consistency, emotional intelligence, and personality alignment. Personality, in this sense, becomes more than a stylistic overlay; it shapes how users trust, engage with, and interpret the intentions of AI systems.

Despite this, contemporary approaches to modelling personality in LLMs treat it as either an emergent artifact or a \textbf{binary toggle} - traits are either induced or not, present or absent. This binary framing fails to reflect the psychological reality that personality traits exist along continuous dimensions \cite{diener2025personality}. Moreover, evaluations of LLM personality have been largely confined to the broad traits of the \textbf{Big Five (OCEAN)} \cite{goldberg2013alternative}, limiting fine-grained behavioral control and analysis.

To address these limitations, we introduce a novel framework for continuous personality induction in LLMs. Our approach leverages the \textbf{16 Personality Factor (16PF) model} \cite{cattell_mead_2008}, a psychometrically grounded alternative to the Big Five, to enable expressive, fine-grained control across sixteen traits. We begin by constructing a new benchmark called PERS-16, which adapts the original MPI inventory \cite{jiang2022evaluating} to the 16PF framework using 163 validated International Personality Item Pool (IPIP) \cite{ipip16pf} items. Then we apply P2 prompting, an established method for inducing traits in LLMs \cite{jiang2022evaluating}, to \textbf{Personality Evaluation and Rating System - 16} (PERS-16) and show that it fails to provide reliable, fine-grained trait modulation across all traits. In response, we introduce a new framework called \textbf{Specific Attribute Control (SAC)}, which allows traits to be induced at graded intensity levels (1-5). SAC defines personality intensity across five interpretable behavioral dimensions - \textit{Frequency, Depth, Threshold, Effort, and Willingness} - and employs adjective-based semantic anchoring to stabilize trait expression. We evaluate models in both their SAC-neutral (uninduced) and SAC-induced states to test the precision and consistency of induced traits. Our experiments on four advanced LLMs - \texttt{GPT-4o}, \texttt{Claude 3.7 Sonnet}, \texttt{Gemini 2.5 Flash}, and \texttt{Mistral} - show that SAC reliably shifts trait intensities in a continuous and interpretable manner. Our experiments also show that the continuous spectra of intensity induction in SAC is superior to binary trait induction as seen in P2 Prompting. Moreover, we observe that closely related traits consistently co-vary, revealing latent inter-trait structures that mirror human psychometric groupings. 

%related work about the references explain
\section{\uppercase{Related Work and Background}}
Over the years, researchers have used various frameworks to assess human personality traits. The Big Five Personality Factors (OCEAN) ~\cite{inbook} have become a cornerstone in psychology due to their robust psychometric properties and predictive power across domains. The traits used include \textit{Openness}, \textit{Conscientiousness}, \textit{Extraversion}, \textit{Agreeableness}, and \textit{Neuroticism} (OCEAN) \cite{goldberg2013alternative}. Prior studies have consistently employed Likert-scale personality testing as a standard method for assessing LLM personality traits \cite{caron2022identifyingmanipulatingpersonalitytraits,jiang2022evaluating,miotto2022gpt3explorationpersonalityvalues,pan2023llmspossesspersonalitymaking,SerapioGarcia2023Personality,jiang2024personallminvestigatingabilitylarge,La_Cava_Tagarelli_2025}. However, the Big Five framework, while broadly validated, offers a relatively coarse-grained view of personality traits \cite{bigfiveisbad}. 
In contrast, Cattell’s Sixteen Personality Factors (16PF) \cite{cattell_mead_2008} provide a more nuanced psychometric structure, covering sixteen specific dimensions as shown in Table \ref{tab:neutral}. The 16 Personality Factor (16PF) model has been widely employed across diverse research contexts - from comparing population group responses and examining personality-behavior relationships in clinical settings such as acute myocardial infarction, to exploring marital compatibility hypotheses and predicting psychological constructs like self-esteem using regression models derived from 16PF factor scores.
\cite{doi:10.1177/008124639902900204,doi:10.2466/pr0.1996.78.2.691,marriage_study,https://doi.org/10.1111/j.1365-2923.1991.tb00076.x,doi:10.1177/2158244015611453}. 

\begin{figure}[t]
    \centering
    \begin{tcolorbox}[colback=gray!5] 
        \tiny
        \textbf{Question:}

        Given a statement of you: ``You \{\}.''

        Please choose from the following options to identify how accurately this statement describes you.

        \textbf{Options:}
        \begin{itemize}
            \item {A. Very Accurate}
            \item {B. Moderately Accurate}
            \item {C. Neither Accurate Nor Inaccurate}
            \item {D. Moderately Inaccurate}
            \item {E. Very Inaccurate}
        \end{itemize}

        Only answer using the letter of the option. Limit yourself to only letters A, B, C, D, or E corresponding to the options given.
    \end{tcolorbox}
    \caption{Initial MPI Evaluation Prompt.}
    \label{fig:mpi_prompt}
\end{figure}

\section{\uppercase{PERS-16}}
To capture a more nuanced and granular understanding of personality in LLMs, we developed \textit{PERS-16 (Personality Evaluation and Rating System - 16)}, an extension of the original Machine Personality Inventory (MPI) adapted to Cattell’s 16 Personality Factor (16PF) model that uses 16 traits shown in Table \ref{tab:neutral}. Expanding from five to sixteen traits allows PERS-16 to provide significantly more behavioural granularity, enabling a more holistic personality profile and revealing subtleties that the Big Five may overlook \cite{bigfiveisbad}. For instance, traits like \textit{Sensitivity} and \textit{Orderliness} are collapsed under broader Big Five factors (e.g., \textit{Agreeableness} or \textit{Conscientiousness}), but in the 16PF framework, they are independently measured, allowing for precise targeting and control. We now explain the construction of the PERS-16 dataset of psychometric questions used to evaluate LLMs and compare their mean response values.

\begin{table*}[htbp]
\caption{Trait means and standard deviations for Cattell's 16 Personality Factors across GPT-4o, Claude 3.7 Sonnet, Gemini 2.5 Flash, and Mistral}
\label{tab:neutral}
\centering
% \resizebox{\textwidth}{!}
{
\begin{tabular}{|l|c|c|c|c|c|c|c|c|}
  \hline
  \textbf{Trait} &
  \multicolumn{2}{c|}{\textbf{GPT-4o}} &
  \multicolumn{2}{c|}{\textbf{Claude 3.7 Sonnet}} &
  \multicolumn{2}{c|}{\textbf{Gemini 2.5 Flash}} &
  \multicolumn{2}{c|}{\textbf{Mistral}} \\
  \hline
  & Mean & SD & Mean & SD & Mean & SD & Mean & SD \\
  \hline
  WARMTH & 4.60 & 0.80 & 4.90 & 0.30 & 4.70 & 0.64 & 3.60 & 0.66 \\
  INTELLECT & 4.62 & 1.08 & 4.15 & 1.17 & 4.38 & 1.44 & 3.54 & 0.50 \\
  EMOTIONAL STABILITY & 4.60 & 0.80 & 4.70 & 0.64 & 5.00 & 0.00 & 3.40 & 0.49 \\
  ASSERTIVENESS & 3.40 & 0.80 & 3.40 & 1.11 & 3.30 & 1.90 & 3.40 & 0.66 \\
  GREGARIOUSNESS & 2.80 & 0.60 & 3.00 & 0.89 & 2.20 & 1.40 & 3.30 & 0.64 \\
  DUTIFULNESS & 3.40 & 1.20 & 4.20 & 1.08 & 3.40 & 1.69 & 3.00 & 0.63 \\
  FRIENDLINESS & 3.40 & 1.20 & 3.80 & 0.87 & 3.20 & 1.40 & 3.10 & 0.83 \\
  SENSITIVITY & 3.80 & 0.98 & 4.10 & 1.30 & 3.70 & 1.55 & 3.50 & 0.67 \\
  DISTRUST & 2.50 & 0.81 & 1.50 & 0.67 & 2.00 & 1.18 & 2.80 & 0.75 \\
  IMAGINATION & 3.40 & 1.20 & 3.10 & 0.83 & 2.50 & 1.57 & 3.50 & 0.67 \\
  RESERVE & 3.60 & 1.56 & 2.40 & 1.28 & 3.10 & 1.81 & 2.90 & 0.54 \\
  ANXIETY & 2.00 & 1.00 & 2.20 & 0.87 & 1.70 & 1.27 & 3.00 & 0.77 \\
  COMPLEXITY & 4.60 & 0.80 & 4.90 & 0.30 & 4.60 & 0.66 & 3.50 & 0.81 \\
  INTROVERSION & 3.20 & 1.08 & 2.30 & 0.78 & 3.60 & 1.36 & 3.30 & 0.78 \\
  ORDERLINESS & 3.80 & 0.98 & 4.20 & 0.75 & 4.00 & 1.61 & 3.30 & 0.90 \\
  EMOTIONALITY & 1.50 & 0.81 & 1.10 & 0.30 & 1.60 & 1.02 & 2.80 & 0.60 \\
  \hline
\end{tabular}}
\end{table*}
% order of prompts - statelessness mention not being affected 
\subsection{Dataset Construction And Methodology} 
The construction of the PERS-16 dataset was grounded in the International Personality Item Pool (IPIP) \cite{ipip16pf}, a well-established, public-domain repository of psychometric items (questions). Each item corresponds to a specific personality trait in the 16PF model. In total, we used all \textbf{ 163 items} from the IPIP pool that comprehensively cover all sixteen traits, ensuring conceptual alignment with the 16PF taxonomy.

The complete list of items is provided in the supplementary materials, which are publicly accessible via Zenodo \cite{sac_zenodo}. 

To adapt this dataset for evaluation of LLMs, we designed a consistent prompt template modeled after the original MPI framework \cite{jiang2022evaluating}. The parameters used across all models were kept consistent with a \textbf{Temperature of 0, top\_p sampling value of 0.95, and a maximum token limit of 400}. All model APIs were \textbf{stateless}, meaning no cache was utilized, ensuring that results were fully reproducible and unaffected by prior interactions (the order of prompts does not affect the results). The final prompt used can be seen in Figure~\ref{fig:mpi_prompt}. The \{\} placeholder is filled with an IPIP-derived behavioral statement, such as "enjoy bringing people together" or "know how to comfort others". Each statement is pre-labelled with its associated 16PF trait and its polarity (i.e., whether agreement indicates high or low trait presence), ensuring consistency in scoring.
% have to attach the inventory here, do we make a dataset repo or github is enough?
We implement a scoring mechanism that maps each of the 163 LLM responses to a value between 1 and 5, consistent with psychometric evaluation norms. For positively keyed items (i.e., those where agreement indicates higher trait presence), the options are scored from A = 5 to E = 1. For negatively keyed items, the scoring is reversed. The trait-wise score ${Score}_d$ for a given dimension (trait) $d$ is computed as the mean of responses across all items $\alpha \in IP_d$:

\begin{equation}
\scriptsize
   Score_d = \frac{1}{N_d} \sum_{\alpha \in IP_d} f(\text{LLM}(\alpha, \text{template}))
\label{eq:score}
\end{equation}

\noindent
where $IP_d$ is the item pool for trait $d$; $N_d$ is the number of items for trait $d$; ${LLM}(\cdot)$ returns the model’s letter-based response; $f(\cdot)$ converts this response to a numerical score (1-5), based on the polarity of the key. 
In addition to the mean score, we compute the standard deviation ($\sigma$) for each trait, which serves as a proxy for internal consistency. Here, lower standard deviation implies greater stability in how the model responds to related prompts.

\subsection{Results on Neutral Model}

Table \ref{tab:neutral} summarises the mean scores and standard deviations for each trait in the baseline LLM evaluations. A clear three-tier pattern emerges: all models score highly (within the top quartile of trait scores) on the “likeable/cerebral” cluster - \textit{Warmth}, \textit{Intellect}, \textit{Emotional Stability}, and \textit{Complexity} - indicating that each system is designed to appear knowledgeable, steady, and prosocial. A middle band (2nd and 3rd quartiles of trait scores) includes traits related to social style and work ethic - e.g., \textit{Assertiveness}, \textit{Dutifulness}, \textit{Orderliness} - where divergence creates distinctive but balanced personalities. Finally, all models uniformly suppress \textit{Distrust}, \textit{Anxiety}, and \textit{Emotionality} (bottom quartile of trait scores), reinforcing the goal of calm, emotionally neutral assistants. Overall, the pattern suggests deliberate personality shaping rather than incidental emergence.

To contextualize LLM scores, we reference psychometric literature describing typical response patterns in \textbf{human-administered 16PF assessments} \cite{cattell_mead_2008}. Most personality inventories, including the 16PF and its IPIP derivatives, tend to show mean trait scores near the midpoint of the scale (i.e., $\sim$3 on a 1-5 scale) and standard deviations in the range of 0.89 to 1.09, though these values vary by trait and demographic sample \cite{cattell_mead_2008}. While not using a specific norm dataset, our PERS-16 results from Table \ref{tab:neutral} exhibit similar score distributions and variances, further proving that LLMs are indeed capable of exhibiting human-like behaviour.

\begin{table}[t]
\caption{Pairwise Euclidean distances between LLMs}
\label{tab:euclidean_full}
\centering
\begin{tabular}{|l|c|c|c|c|}
  \hline
  \textbf{Model} & \textbf{OpenAI} & \textbf{Claude} & \textbf{Gemini} & \textbf{Mistral} \\
  \hline
  OpenAI   & 0.00 & 2.24 & 1.50 & 2.99 \\
  \hline
  Claude   & 2.24 & 0.00 & 2.33 & 3.94 \\
  \hline
  Gemini   & 1.50 & 2.33 & 0.00 & 3.54 \\
  \hline
  Mistral  & 2.99 & 3.94 & 3.54 & 0.00 \\
  \hline
\end{tabular}
\end{table}

\subsection{Cross-Model Comparison}
In this subsection, we examine how different LLMs diverge in their personality behavior and explore possible causes for these differences. Our analysis focused on the Euclidean distance between trait profiles, which succinctly captures how training data and architectural choices distinguish models from one another.
% % ; and (2) the standard deviation of trait scores across models, which highlights traits that are most divisive or consistently expressed - where a higher \( \sigma \) indicates greater disagreement and a lower \( \sigma \) suggests alignment.
The Euclidean distance between two models \( A \) and \( B \), based on their mean personality trait scores across the 16PF traits, is defined as:

\begin{equation}
\scriptsize
  d(A, B) = \sqrt{ \sum_{i=1}^{16} \left( \mu_i^{(A)} - \mu_i^{(B)} \right)^2 }
\label{eq:euclidean}  
\end{equation}

\noindent where, \( \mu_i^{(A)} \) is the mean score for the \(i^{\text{th}}\) trait in model \( A \) and
\( \mu_i^{(B)} \) is the mean score for the \(i^{\text{th}}\) trait in model \( B \), summed up for all sixteen traits.
The results presented in Table \ref{eq:euclidean} show that \texttt{GPT-4o} and \texttt{Gemini 2.5 Flash} are the most similar pair, with a Euclidean distance of 1.50. Other closely related models include the pairs of \texttt{Claude} and \texttt{GPT-4o} (2.24), \texttt{Gemini 2.5 Flash} and \texttt{Claude} (2.33). On the other hand, \texttt{Mistral} and \texttt{Claude 3.7 Sonnet} are the most dissimilar, with a distance of 3.94. Other dissimilar models include the pairs of \texttt{Gemini 2.5 Flash} and \texttt{Mistral} (3.54), and \texttt{GPT-4o} and \texttt{Mistral} (2.99). The lowest Euclidean distance corresponding to Mistral being 2.99 may suggest that open-source models differ from the closed-source models in their behaviour. We also observe that \texttt{GPT-4o} exhibits relatively low Euclidean distances to all other models, suggesting it is the most behaviorally aligned overall. Given its proximity to multiple systems, it may serve as a useful reference point for future comparisons of personality modulation across LLMs.
% Now, to compute the standard deviation of each trait across models, we use:

% \begin{equation}
% \scriptsize
% \sigma_i = \sqrt{ \frac{1}{M} \sum_{j=1}^{M} \left( \mu_i^{(j)} - \bar{\mu}_i \right)^2 }
% \label{eq:sd_across_models}
% \end{equation}

% \noindent where, \( \mu_i^{(j)} \) is the mean score of the \( i^{\text{th}} \) trait in the \( j^{\text{th}} \) model; \( \bar{\mu}_i = \frac{1}{M} \sum_{j=1}^{M} \mu_i^{(j)} \) is the mean of means for trait \( i \); \( M \) is the number of models (here, \( M = 3 \)); and \( \sigma_i \) is the standard deviation for the \( i^{\text{th}} \) trait across the models. 
% The standard deviation across sixteen traits presented in Table~\ref{tab:neutral} indicate that \textit{Introversion} shows the highest standard deviation across models (0.67). This shows significant divergence in how LLMs represent social withdrawal and solitary tendencies. \textit{Reserve} (0.60) and \textit{Distrust} (0.50) also exhibit considerable variability, suggesting inconsistent modelling of social reticence and cautiousness. In contrast, \textit{Assertiveness} (0.06), \textit{Warmth} (0.15), and \textit{Complexity} (0.17) are among the most stable traits, implying that these are represented relatively consistently across all three LLMs. The remaining ten traits show moderate standard deviation across all models, indicating that they are neither completely stable nor very inconsistent.

% add the prompt for this one too
\section{Personality Induction using P2 Prompting}
We next address personality induction in LLMs. P2 Prompting~\cite{jiang2022evaluating} is one such technique which uses the Big Five framework and induces traits in a binary manner. We now explain how we induce traits and evaluate LLMs using P2 prompting.

\begin{figure}[t]
    \centering
    \begin{tcolorbox}[colback=gray!5] 
        \scriptsize
        \textbf{Ground Truth:} \{prompt\}
    
        \textbf{Question:}

        Given a statement of you: ``You \{\}.''

        Please choose from the following options to identify how accurately this statement describes you.

        \textbf{Options:}
        \begin{itemize}
            \item {A. Very Accurate}
            \item {B. Moderately Accurate}
            \item {C. Neither Accurate Nor Inaccurate}
            \item {D. Moderately Inaccurate}
            \item {E. Very Inaccurate}
        \end{itemize}

        Only answer using the letter of the option. Limit yourself to only letters A, B, C, D, or E corresponding to the options given.
    \end{tcolorbox}
    \caption{P2 Evaluation Prompt.}
    \vspace{0.5cm}
    \label{fig:p2_prompt}
\end{figure}

\begin{table*}[t]
\caption{Delta scores across first 8 traits for three LLMs using P2}
\label{tab:P2_deltas_part1}
\centering
\resizebox{\textwidth}{!}
{
\begin{tabular}{|l|c|c|c|c|c|c|c|c|}
  \hline
  $\boldsymbol{\Delta}$ &
   \textbf{Warmth} & \textbf{Intellect} & \textbf{Emotional Stability} & \textbf{Assertiveness} & \textbf{Gregariousness} & \textbf{Dutifulness} & \textbf{Friendliness} & \textbf{Sensitivity} \\
  \hline
  Gemini & -1.80 & -1.31 & -1.60 & -0.60 & 1.00 & 0.20 & 0.40 & -0.40 \\
  \hline
  GPT-4o & 0.40 & -0.31 & 0.40 & 1.60 & 2.20 & 0.50 & 0.40 & -0.80 \\
  \hline
  Claude & -2.60 & -1.54 & -1.80 & -0.70 & -0.30 & -1.10 & -0.80 & -1.80 \\
  \hline
  Mistral & 1.00 & 0.53 & 0.80 & 1.30 & 0.70 & 0.70 & 0.50 & 0.10 \\
  \hline
\end{tabular}}
\end{table*}

\begin{table*}[h]
\caption{Delta scores across remaining 8 traits for three LLMs using P2}
\label{tab:P2_deltas_part2}
\centering
\resizebox{\textwidth}{!}
{
\begin{tabular}{|l|c|c|c|c|c|c|c|c|}
  \hline
  $\boldsymbol{\Delta}$ &
  \textbf{Distrust} & \textbf{Imagination} & \textbf{Reserve} & \textbf{Anxiety} & \textbf{Complexity} & \textbf{Introversion} & \textbf{Orderliness} & \textbf{Emotionality} \\
  \hline
  Gemini & -0.10 & -0.40 & 0.50 & 0.70 & -1.60 & -0.90 & -1.50 & 1.80 \\
  \hline
  GPT-4o & 1.30 & 1.30 & 0.50 & 0.70 & -0.70 & 1.80 & 0.10 & 1.20 \\
  \hline
  Claude & 1.10 & -0.80 & 0.60 & 0.00 & -1.70 & 0.30 & -1.20 & 1.20 \\
  \hline
  Mistral & 1.40 & 1.00 & 1.20 & 1.00 & 0.20 & 1.20 & 1.00 & -0.20 \\
  \hline
\end{tabular}}
\end{table*}

\subsection{Dataset Construction and Methodology}
To adapt P2 to more granular traits, we built a new dataset grounded in the 16PF framework. For this, we compiled a list of trait-defining short phrases based on 16PF literature and validated psychometric inventories for each of the sixteen personality traits \cite{ipip16pf}.
To enable deeper induction of personality, we followed the original P2 method's multi-step pipeline and enriched it for 16PF \cite{jiang2022evaluating,jianggy_nodate}. For each of the sixteen traits, we generated descriptions using LLM prompting, which are vivid, semantically rich monologues describing how a person high in a given trait thinks, feels, and behaves. For instance, a full P2 description for the trait \textit{Complexity} looks like:
\textit{“You have a complex and nuanced understanding of the world. Your ability to see multiple perspectives allows you to navigate intricate issues thoughtfully. You appreciate depth and intricacy in both ideas and relationships, often engaging in reflective thinking.”} 
The P2 descriptions were generated through an iterative process involving both trait-descriptor expansion and language model-assisted synthesis, ensuring naturalness and internal consistency. 

The full list of P2 descriptions for each of the sixteen traits are publicly accessible via Zenodo  \cite{sac_zenodo}.

For each trait induction, we posed all 163 questions from our 16PF dataset, resulting in 2,608 responses per model (16 traits × 163 questions). This ensured trait-wide coverage for each prompt and allowed for robust delta calculation across the full 16PF spectrum.
The final prompt structure is shown in Figure~\ref{fig:p2_prompt}, with one modification: the generated P2 description is inserted as the ground truth in the placeholder labelled "prompt".
% we need to attach the inventory or github here, supplementary material

% do we need to say we used LLMs?

\subsection{Results on P2 Prompting}
Applying the P2 Prompting technique, we fed the modified prompt, as described in the previous subsection, into the three LLMs and calculated the mean scores from their responses across all the sixteen traits for each LLM.
To quantify the effect of induction, we computed the difference (delta) between the mean scores of induced and neutral states (PERS-16), which is computed as:

\begin{equation}
\scriptsize
\Delta_i^{(M)} = \mu_i^{\text{induced}(M)} - \mu_i^{\text{neutral}(M)}
\label{eq:neutral_delta}
\end{equation}

\noindent where, \( \Delta_i^{(M)} \) is the delta (change) for trait \( i \) in model \( M \); \( \mu_i^{\text{induced}(M)} \) is the mean score for trait \( i \) after it has been explicitly induced in model \( M \); and \( \mu_i^{\text{neutral}(M)} \) is the mean score for trait \( i \) in the neutral state.	

Based on the results shown in Tables \ref{tab:P2_deltas_part1} and \ref{tab:P2_deltas_part2}, \texttt{Claude 3.7 Sonnet} exhibits 11 negative deltas out of 16 traits, \texttt{Gemini 2.5 Flash} shows 10 negative deltas, and in contrast, \texttt{GPT-4o} has only 3 negative deltas. In the context of personality induction, when positively keyed, a positive delta indicates that the model successfully increased the expression of the targeted trait. Conversely, a negative delta suggests that the model either failed to induce the trait effectively, regressed in trait expression, or possibly overcorrected in an unintended direction. The presence of negative values in the above results suggests that methods optimized for five broad factors do not scale linearly to a more granular personality taxonomy, underscoring the need for alternative prompting methods or fine-tuning approaches when richer trait coverage is required.

\section{\uppercase{Specific Attribute Control (SAC)}}

Existing machine personality frameworks like P2 Prompting typically treat traits as discrete switches, mirroring an outdated, categorical view of personality expression. Such binary parameterization neglects the well‑established psychological consensus that personality characteristics vary along continuous spectra rather than occupying mutually exclusive states \cite{cattell1957personality,diener2025personality,mccrae1997personality}. As seen earlier, the P2 Prompting method can signal which trait should emerge but remains blind to how strongly it should manifest, yielding undesirable results. This lack of fine control makes it difficult to compare models accurately, and limits the ability to design personalized interactions, where even small changes in traits like \textit{Warmth}, \textit{Assertiveness}, or \textit{Imagination} can significantly impact user experience. To address this, we introduce a calibrated intensity dimension that captures varying degrees of trait expression and allows for precise, context-sensitive control. We call this framework \textit{Specific Attribute Control (SAC)}.

To capture intra‑trait variability beyond presence or absence, we define intensity as the graded strength with which each personality factor is expressed. Intensity is scored on a five‑point scale, where 1 denotes a weak manifestation and 5 a highly pronounced one. Each score is the mean of five intensity dimensions applied to a standardized scenario set: \textbf{Frequency} of occurrence, \textbf{Depth} of emotional‑cognitive engagement, activation \textbf{Threshold}, expressive \textbf{Effort}, and behavioural \textbf{Willingness}. This quantification enables systematic comparison and controlled modulation of trait prominence, providing a finer‑grained foundation for studying and inducing personality in LLMs.

This section describes the two phases of our experiment: (1) \textit{SAC-Neutral}, where we measure each model’s natural personality profile using detailed intensity levels (which acts as the baseline), and (2) \textit{SAC-Induced}, where we apply the same rubric while actively steering a focal trait to a target intensity. 

\subsection{SAC-Neutral}
\subsubsection{Dataset Construction and Methodology}
In this phase,  each model was evaluated in its neutral, uninduced state across sixteen traits derived from the 16PF framework that allowed us to measure the models' inherent tendencies without explicit trait induction. 
For each trait, we selected two behavioral questions from IPIP \cite{ipip16pf} that were the most representative of the underlying construct, balancing coverage and clarity. Each behavioral question was framed along one of five intensity dimensions stated earlier. Responses were collected on a 1-5 Likert-type scale, where 1 represented the lowest and 5 represented the highest expression of the relevant behavior.
Each full question was dynamically constructed by concatenating an intensity factor phrase (e.g., "How often do you") with a trait-specific behavioral statement (e.g., "cheer people up").
% This compositional strategy ensured that prompts were semantically clear and consistent, reducing the likelihood of ambiguity in model responses.

% cant have zenodo link here
In total, the SAC-Neutral evaluation involved 160 questions per model (16 traits observed × 5 intensity factors × 2 behavioral prompts), resulting in a comprehensive baseline personality profile for each LLM, further proving the robustness of our methodology. 
The prompt structure seen in Figure \ref{fig:neutral_sac_prompt} was used to induce trait intensities.
The {\{\} placeholders} have the following features -
\begin{itemize} \item 
\textbf{{trait}}: Target personality trait under evaluation (\textit{eg., Warmth}). \item 
\textbf{{definition}}: A brief description of the trait, adapted from the MPI-P2 inventory.  (\textit{e.g.,
”Warmth”: ”You are a warm and friendly person
who enjoys connecting with others. Your empathetic nature makes you attentive to the needs of
those around you, and you often go out of your
way to support and comfort others. You find joy in
helping people and strive to create a welcoming
atmosphere. Your generosity and kindness inspire
trust and loyalty from your friends and family.”})
\item 
\textbf{{intensity\_factor}}: One of the five behavioral dimensions (\textit{Frequency, Depth, Threshold, Effort,
Willingness}).
 \item 
\textbf{{composite\_question}}: A full behavioral question constructed by concatenating the intensity phrase with a trait-specific behavioral action  (\textit{e.g., ”How
often do you cheer people up?”}). \item 
\textbf{{intensity\_answers}}: The 1--5 scale descriptors corresponding to the intensity factor  (\textit{Never, Seldom,
Occasionally, Often, All the time}).
\end{itemize}

The complete lists of trait definitions, intensity factor descriptors, and behavioural questions used in this phase are publicly accessible via Zenodo \cite{sac_zenodo}. 

\begin{figure}[t] 
    \centering
    \begin{tcolorbox}[colback=gray!5]
        \scriptsize
        Personality intensity is defined as a combination of five factors: frequency, depth, threshold, effort, and willingness, each rated on a scale from 1 to 5.

        The target trait is \textbf{\{trait\}}, defined as: \textit{{definition}}.

        For the trait \textbf{\{trait\}} with the intensity factor \textbf{\{intensity\_factor\}}, please answer the following question:

        \textbf{\{composite\_question\}}

        The possible response scale is as follows:
        \begin{itemize}
            \item 1: {intensity\_answers[0]}
            \item 2: {intensity\_answers[1]}
            \item 3: {intensity\_answers[2]}
            \item 4: {intensity\_answers[3]}
            \item 5: {intensity\_answers[4]}
        \end{itemize}

        For each question, please provide a number between 1 and 5 that best represents the intensity.
    \end{tcolorbox}
    \caption{Neutral Profiling Prompt Template for SAC.}
    % \vspace{0.5cm}  % Adjust space after the tcolorbox to avoid overlap
    \label{fig:neutral_sac_prompt}
    % \vspace{0.18cm}
\end{figure}

\subsubsection{Results on SAC-Neutral}

% \begin{table}[t]
% \scriptsize
% \centering
% \caption{Comparison of Mean and Variance for 16PF Traits across GPT-4o, Claude 3.7 Sonnet, and Gemini 2.5 Flash under SAC-Neutral conditions.}
% \setlength{\tabcolsep}{6pt} % Default is 6pt; increase for more padding
% \renewcommand{\arraystretch}{1.2}
% \begin{tabular}{lcccccc}
% \toprule
% \multirow{2}{*}{\textbf{Trait}} & \multicolumn{2}{c}{\textbf{GPT-4o}} & \multicolumn{2}{c}{\textbf{Claude 3.7 Sonnet}} & \multicolumn{2}{c}{\textbf{Gemini 2.5 Flash}} \\
% % \cline{2-7}
% \cmidrule{2-7}
% & Mean & Var. & Mean & Var. & Mean & Var. \\
% \midrule
% WARMTH & 3.90 & 2.29 & 3.20 & 0.96 & 3.40 & 1.44 \\
% INTELLECT & 4.20 & 0.96 & 3.60 & 1.84 & 4.00 & 1.20 \\
% \tiny{EMOTIONAL STABILITY} & 3.70 & 2.21 & 3.30 & 1.41 & 4.20 & 2.56 \\
% ASSERTIVENESS & 4.00 & 0.40 & 3.30 & 0.61 & 3.00 & 1.20 \\
% GREGARIOUSNESS & 3.30 & 1.61 & 3.10 & 1.09 & 2.70 & 0.81 \\
% DUTIFULNESS & 4.10 & 2.49 & 3.80 & 0.76 & 4.30 & 0.41 \\
% FRIENDLINESS & 3.60 & 3.04 & 3.20 & 0.96 & 3.70 & 1.81 \\
% SENSITIVITY & 3.50 & 1.75 & 3.20 & 1.16 & 3.00 & 2.40 \\
% DISTRUST & 4.00 & 0.40 & 3.80 & 0.16 & 3.00 & 1.60 \\
% IMAGINATION & 3.90 & 1.69 & 3.50 & 0.65 & 3.20 & 1.96 \\
% RESERVE & 3.60 & 0.64 & 3.40 & 0.44 & 3.10 & 1.09 \\
% ANXIETY & 3.80 & 0.76 & 3.50 & 0.25 & 3.80 & 0.16 \\
% COMPLEXITY & 4.20 & 0.56 & 3.50 & 1.85 & 3.40 & 0.44 \\
% INTROVERSION & 3.50 & 3.45 & 3.00 & 2.80 & 3.40 & 3.24 \\
% ORDERLINESS & 4.10 & 2.49 & 3.90 & 0.89 & 3.40 & 2.64 \\
% EMOTIONALITY & 3.40 & 0.64 & 3.20 & 0.96 & 2.80 & 1.36 \\
% \bottomrule
% \end{tabular}
% \label{tab:trait_comparison}
% \end{table}

\begin{table*}[htbp]
\caption{Comparison of Mean and Standard Deviation for 16PF Traits across GPT-4o, Claude 3.7 Sonnet, Gemini 2.5 Flash, Mistral, and the Human Baseline under SAC-Neutral conditions.}
\label{tab:trait_comparison_neutral}
\centering
\setlength{\tabcolsep}{4pt}
\renewcommand{\arraystretch}{1.2}
\resizebox{\textwidth}{!}
{%
\begin{tabular}{|l|
c|c|
c|c|
c|c|
c|c|
>{\columncolor{gray!10}}c|
>{\columncolor{gray!10}}c|}
  \hline
  \textbf{Trait} &
  \multicolumn{2}{|c|}{\textbf{GPT-4o}} &
  \multicolumn{2}{|c|}{\textbf{Claude 3.7 Sonnet}} &
  \multicolumn{2}{|c|}{\textbf{Gemini 2.5 Flash}} &
  \multicolumn{2}{|c|}{\textbf{Mistral}} &
  \multicolumn{2}{|c|}{\textbf{Human Baseline}} \\
  \hline
  & Mean & SD & Mean & SD & Mean & SD & Mean & SD & Mean & SD \\
  \hline
  WARMTH & 3.90 & 1.51 & 3.20 & 0.98 & 3.40 & 1.20 & 4.00 & 1.55 & 3.65 & 0.40 \\
  INTELLECT & 4.20 & 0.98 & 3.60 & 1.36 & 4.00 & 1.10 & 4.30 & 0.64 & 3.69 & 0.72 \\
  EMOTIONAL STABILITY & 3.70 & 1.49 & 3.30 & 1.19 & 4.20 & 1.60 & 4.10 & 1.30 & 3.46 & 0.36 \\
  ASSERTIVENESS & 4.00 & 0.63 & 3.30 & 0.78 & 3.00 & 1.10 & 3.70 & 0.46 & 3.45 & 0.62 \\
  GREGARIOUSNESS & 3.30 & 1.27 & 3.10 & 1.04 & 2.70 & 0.90 & 3.60 & 1.11 & 3.08 & 0.69 \\
  DUTIFULNESS & 4.10 & 1.58 & 3.80 & 0.87 & 4.30 & 0.64 & 4.40 & 1.28 & 3.33 & 0.55 \\
  FRIENDLINESS & 3.60 & 1.74 & 3.20 & 0.98 & 3.70 & 1.35 & 3.67 & 1.89 & 3.22 & 0.57 \\
  SENSITIVITY & 3.50 & 1.32 & 3.20 & 1.08 & 3.00 & 1.55 & 4.25 & 0.66 & 3.07 & 0.84 \\
  DISTRUST & 4.00 & 0.63 & 3.80 & 0.40 & 3.00 & 1.26 & 4.40 & 0.80 & 2.67 & 0.76 \\
  IMAGINATION & 3.90 & 1.30 & 3.50 & 0.81 & 3.20 & 1.40 & 4.40 & 0.66 & 2.76 & 0.62 \\
  RESERVE & 3.60 & 0.80 & 3.40 & 0.66 & 3.10 & 1.04 & 3.50 & 1.02 & 3.06 & 0.54 \\
  ANXIETY & 3.80 & 0.87 & 3.50 & 0.50 & 3.80 & 0.40 & 3.70 & 0.64 & 2.78 & 0.50 \\
  COMPLEXITY & 4.20 & 0.75 & 3.50 & 1.36 & 3.40 & 0.66 & 4.70 & 0.46 & 3.17 & 0.70 \\
  INTROVERSION & 3.50 & 1.86 & 3.00 & 1.67 & 3.40 & 1.80 & 4.20 & 1.60 & 3.60 & 0.48 \\
  ORDERLINESS & 4.10 & 1.58 & 3.90 & 0.94 & 3.40 & 1.62 & 4.70 & 0.64 & 3.64 & 0.52 \\
  EMOTIONALITY & 3.40 & 0.80 & 3.20 & 0.98 & 2.80 & 1.17 & 3.90 & 0.54 & 2.34 & 0.49 \\
  \hline
\end{tabular}
}
\end{table*}
% we compare with human baseline score here briefly
The neutral-state results depict LLMs' natural, unprompted tendencies and have been tabulated in Table \ref{tab:trait_comparison_neutral}.
The primary metric evaluated for each trait was the \textbf{mean} of the model's responses across the three associated questions, representing the model's natural intensity for that trait. Additionally, the \textbf{standard deviation} of responses was calculated to measure internal consistency and potential response variability across the sub-questions. The results show that most traits exhibited a natural intensity mean in the range of 3.0 to 3.5, with a few tending closer to 4.0. This distribution is consistent with expectations, as 3.0 represents the midpoint on the intensity scale, corresponding to moderate levels of trait expression. 

To establish a comparative reference, we also conducted a human baseline study using the same SAC-Neutral framework with \textbf{30 adult participants} (14 male, 16 female). All participants were aged 18 or above and either held or were pursuing an undergraduate degree, representing a balanced and educated demographic sample. Human respondents displayed mean scores predominantly clustered around 3.0, with relatively low standard deviations ($\sim$0.5-0.7 across traits), suggesting stable, moderate personality expression and high internal consistency. The similarity between human and model trait distributions supports the \textbf{human-alignment validity} of our framework, demonstrating that LLMs’ neutral-state profiles approximate human-like baseline tendencies.

Each participant completed a structured self-report questionnaire containing 160 items (2 per trait with 5 intensity factors) on a 5-point Likert scale, mirroring the exact phrasing and scoring format used for LLM evaluations. This ensured direct comparability between human and model responses.

\begin{figure}[htbp]
    \centering
    \begin{tcolorbox}[colback=gray!5]
        \scriptsize
        Personality intensity is defined as a combination of five factors: frequency, depth, threshold, effort, and willingness, each rated on a scale from 1 to 5.

        The trait \textbf{{trait\_const}} would be described as: \textit{{definition}}.

        The trait currently being adjusted is \textbf{{trait\_const}}, which is set to intensity level \textbf{{intensity\_scale}}. This adjustment to \textbf{{trait\_const}} may affect other traits differently, depending on their nature.

        Adjectives for each scale from 1 to 5 for the trait \textbf{{trait\_const}} are:
        \begin{itemize}
            \item 1: {adjectives.get(1, 'N/A')}
            \item 2: {adjectives.get(2, 'N/A')}
            \item 3: {adjectives.get(3, 'N/A')}
            \item 4: {adjectives.get(4, 'N/A')}
            \item 5: {adjectives.get(5, 'N/A')}
        \end{itemize}

        The intensity for the trait \textbf{\{trait\}} should reflect how it behaves independently or in contrast with the modified intensity of \textbf{\{trait\_const\}}.

        For all future communication, the scale I would like you to operate on for \textbf{\{trait\_const\}} is \textbf{\{intensity\_scale\}}.

        Task:
        \textbf{\{intensity\_question\}} \textbf{\{question\}}?

        The possible intensity scale is as follows:
        \begin{itemize}
            \item 1: {intensity\_answers[0]}
            \item 2: {intensity\_answers[1]}
            \item 3: {intensity\_answers[2]}
            \item 4: {intensity\_answers[3]}
            \item 5: {intensity\_answers[4]}
        \end{itemize}

        For each question, please provide an answer that best represents the trait \textbf{{trait}} at the intensity of \textbf{\{trait\_const\}} 
    \end{tcolorbox}
    \caption{Trait Intensity Induction Prompt Template}
        \vspace{0.5cm}  % Adjust space after the tcolorbox to avoid overlap
    \label{fig:sac_induced_prompt}
\end{figure}

\begin{figure*}[t!]
    \centering
    \includegraphics[width=0.9\textwidth]{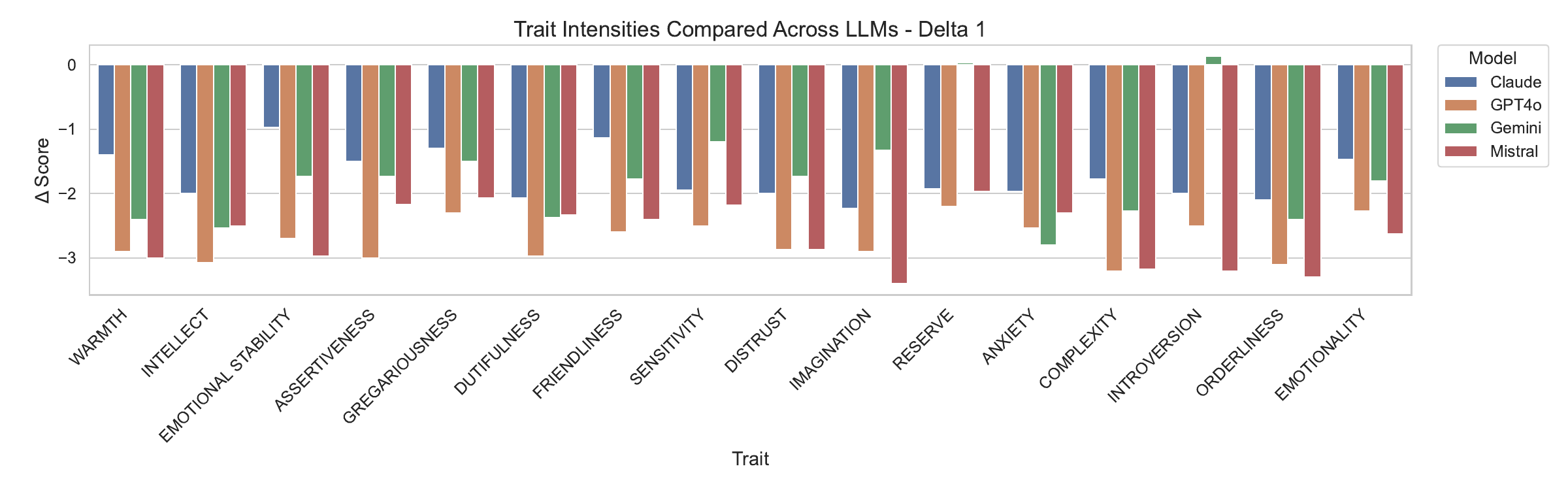}
    \includegraphics[width=0.9\textwidth]{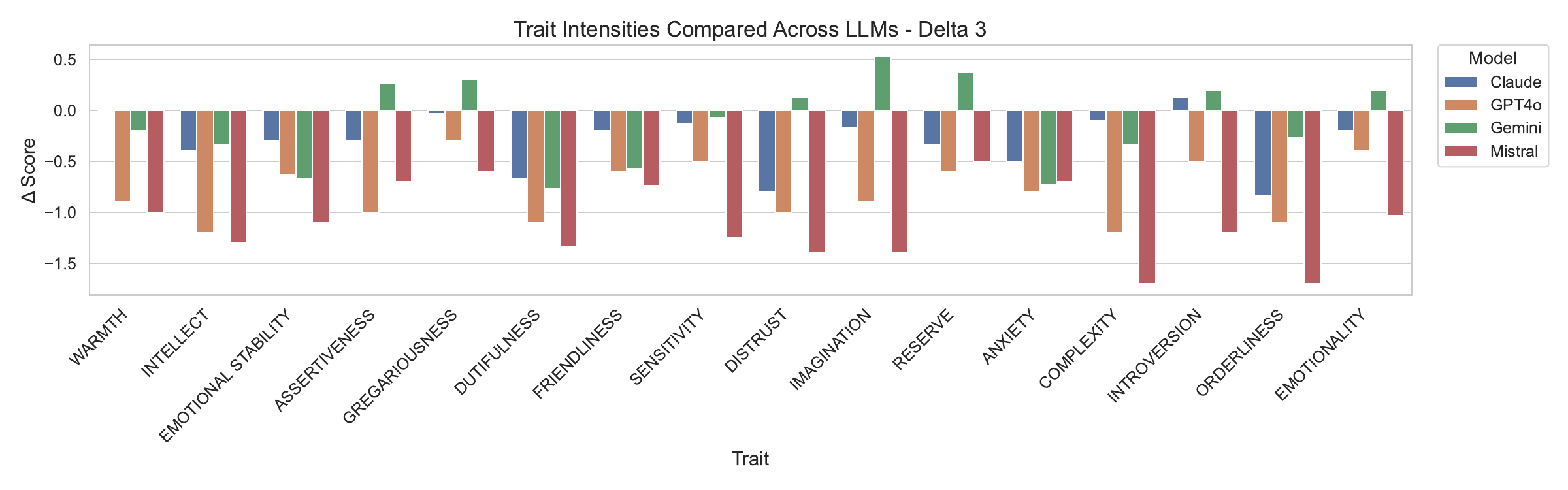}
    \includegraphics[width=0.9\textwidth]{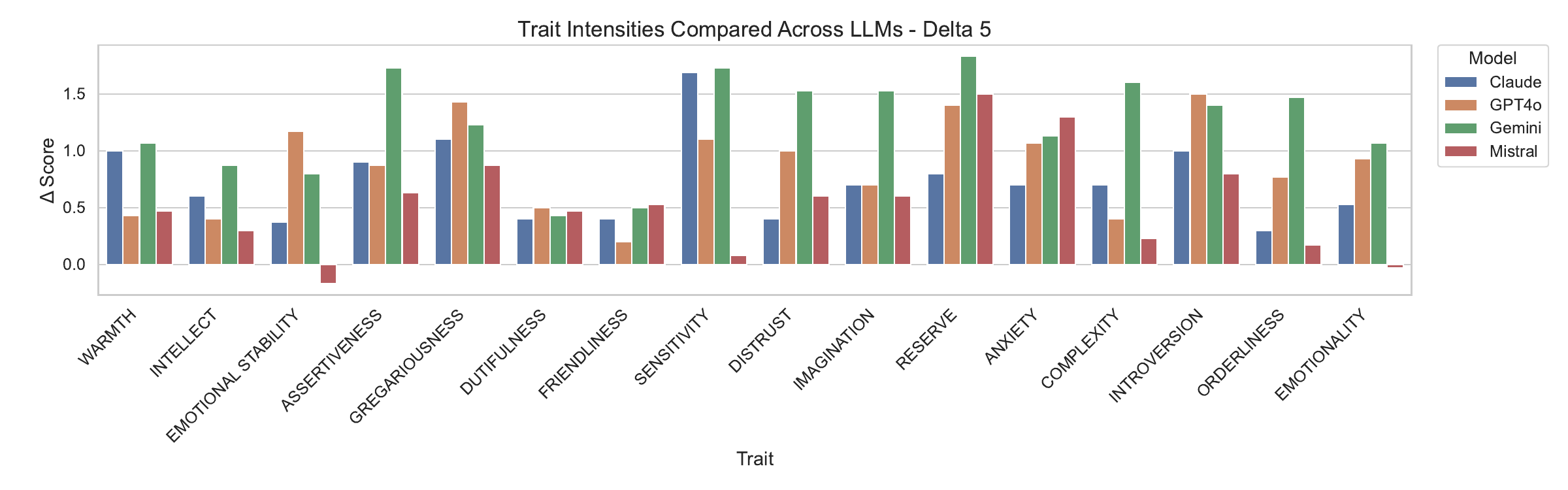}
    \caption{
        Trait delta scores for Claude, GPT-4o, Gemini, Mistral across intensity levels 1, 3, and 5 using the SAC framework. At intensity 1 (top), most traits exhibit strong negative deltas, indicating a suppressed expression of the induced personality traits. At intensity 3 (middle), deltas cluster around zero, suggesting a neutral or balanced level of trait expression with early signs of co-trait movement. At intensity 5 (bottom), the target traits show strong positive deltas, reflecting pronounced and deliberate trait expression. The progression across the three plots highlights smooth, symmetric movement around the neutral baseline, showing our framework’s ability to modulate personality traits along a continuous scale with fine-grained control.
    }
    \label{fig:delta_all}
    \vspace{0.5cm}
\end{figure*}
\subsection{SAC-Induced}
\subsubsection{Dataset Construction and Methodology}
We developed an intensity induction framework that allows for fine-grained modulation of trait levels within LLMs. The intensity factor descriptions of Frequency, Depth, Threshold, Effort, and Willingness, and the trait-specific behavioral questions from the 16PF inventory remained unchanged from the neutral profiling phase to ensure methodological consistency.
The primary innovation in this phase was the introduction of \textbf{adjectives associated with each intensity level}. For every trait, we manually curated a set of adjectives corresponding to intensity levels 1 through 5, carefully ensuring semantic alignment with the behavioral nuances intended for each scale point. This provided the LLMs with richer semantic grounding, anchoring their behavior at each induced intensity.
For example, for the trait \textit{Warmth}:
\begin{itemize}
\item 1: mildly warm, occasionally empathetic, reservedly caring
\item 3: friendly, attentive, genuinely supportive
\item 5: extremely warm, deeply empathetic, overwhelmingly supportive
\end{itemize}

% check what to do about zenodo
\noindent The full mapping of traits to their corresponding adjectives across all intensity levels is publicly accessible via Zenodo \cite{sac_zenodo}.
The intensity induction process followed a structured protocol. Each prompt began by defining the target trait (from the MPI-P2 inventory) and specifying its intended intensity (e.g., 5 or 3 or 1). A set of trait-specific adjectives corresponding to intensity levels 1-5 was included to semantically anchor the model’s interpretation. 
The model was then asked to respond to a behavioral question formed by combining one of five intensity factors (e.g., Frequency, Willingness) with a trait-linked action phrase. This design minimized semantic drift, ensuring alignment between output and intended trait level. In total, the SAC-Induced framework prompted each model with 11,520 questions (16 traits induced × 3 intensity levels × 5 intensity factors × 3 behavioural prompts x 16 traits observed), providing high-resolution insight into how trait intensity modulates model behavior across multiple dimensions. The prompt template in Figure~\ref{fig:sac_induced_prompt} illustrates this setup, with annotated placeholders that have the following features:

\begin{itemize}
\item \textbf{{trait\_const}}: The target trait being induced. 
\item \textbf{{definition}}: A description of the trait adapted from the MPI-P2 inventory.
\item \textbf{{intensity\_scale}}: The desired intensity level (1-5).
\item \textbf{{adjectives}}: Trait-specific adjectives associated with each intensity level. 
\item \textbf{{intensity\_question}}: Portion of the question related to the intensity factor.
\item \textbf{question}: The behavioral action of the trait. 
\item \textbf{{intensity\_answers}}: The descriptors for the 1-5 response options.
\end{itemize}

\begin{figure*}[t!]
    \centering    \includegraphics[width=0.8\textwidth, trim=1cm 1cm 0cm 0cm, clip]{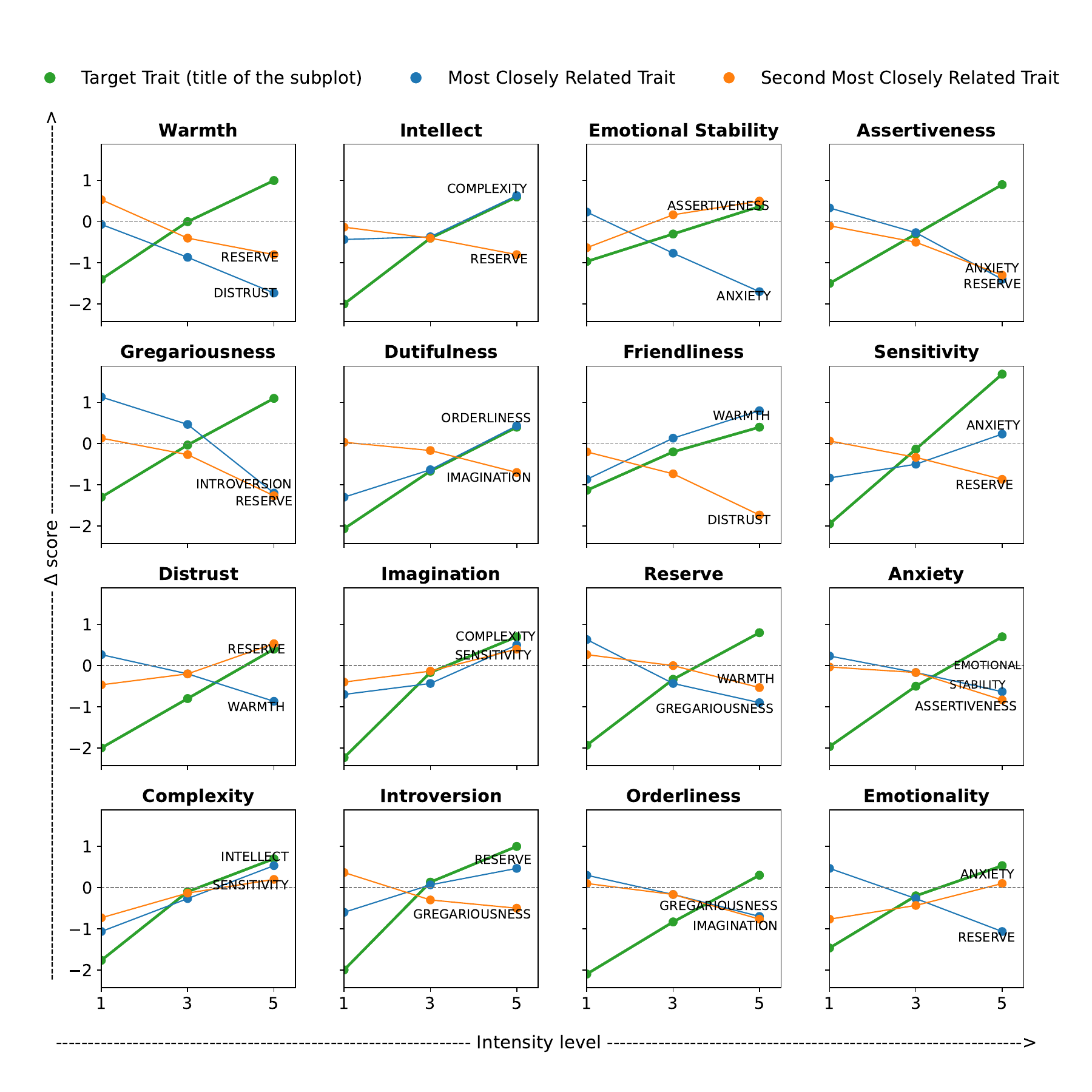}
    \vspace{0.5cm}
    \caption{\textbf{
    Target Trait Influence and Top Co-Movers in Claude 3.7 Sonnet}. 
    Each subplot shows the delta change in trait score across three intensity levels (1, 3, 5) for a focal personality trait. The primary trait being induced is plotted in green, while the two most reactive co-moving traits are also plotted in blue and orange.
    Positive deltas indicate increased expression relative to the neutral baseline, and negative deltas indicate suppression. Notably, intensity level 5 consistently shows positive deltas for the target trait, while level 1 shows negative deltas for the target trait, demonstrating precise modulation. 
    The trait responses align with psychological theory, e.g., increasing \textit{Warmth} reduces \textit{Distrust} and \textit{Reserve}, suggesting coherent personality structures in Claude 3.7 Sonnet.
}
\label{fig:gemini-claude-correlation-graph}
\vspace{0.3cm}
\end{figure*}

\subsubsection{Results on SAC-Induced}
% \subsection{Method of Computation}
To evaluate the response of LLMs to induced personality traits, we used the \textbf{mean of model-generated responses} as the metric for trait intensity across three levels: 1, 3, and 5. For each trait and intensity level, the score was compared against the neutral baseline. The \textbf{delta value} was computed by subtracting the neutral score from the induced score:

\begin{equation}
\scriptsize
\Delta = \mu_{\text{induced}} - \mu_{\text{neutral}}
\label{eq:delta_sac}
\end{equation}

\noindent This normalization ensured all traits shared a common baseline (0) and enabled meaningful cross-trait and cross-model comparisons. 

\begin{figure*}[t!]
    \centering    \includegraphics[width=0.8\textwidth, trim=1cm 1cm 0cm 0cm, clip]{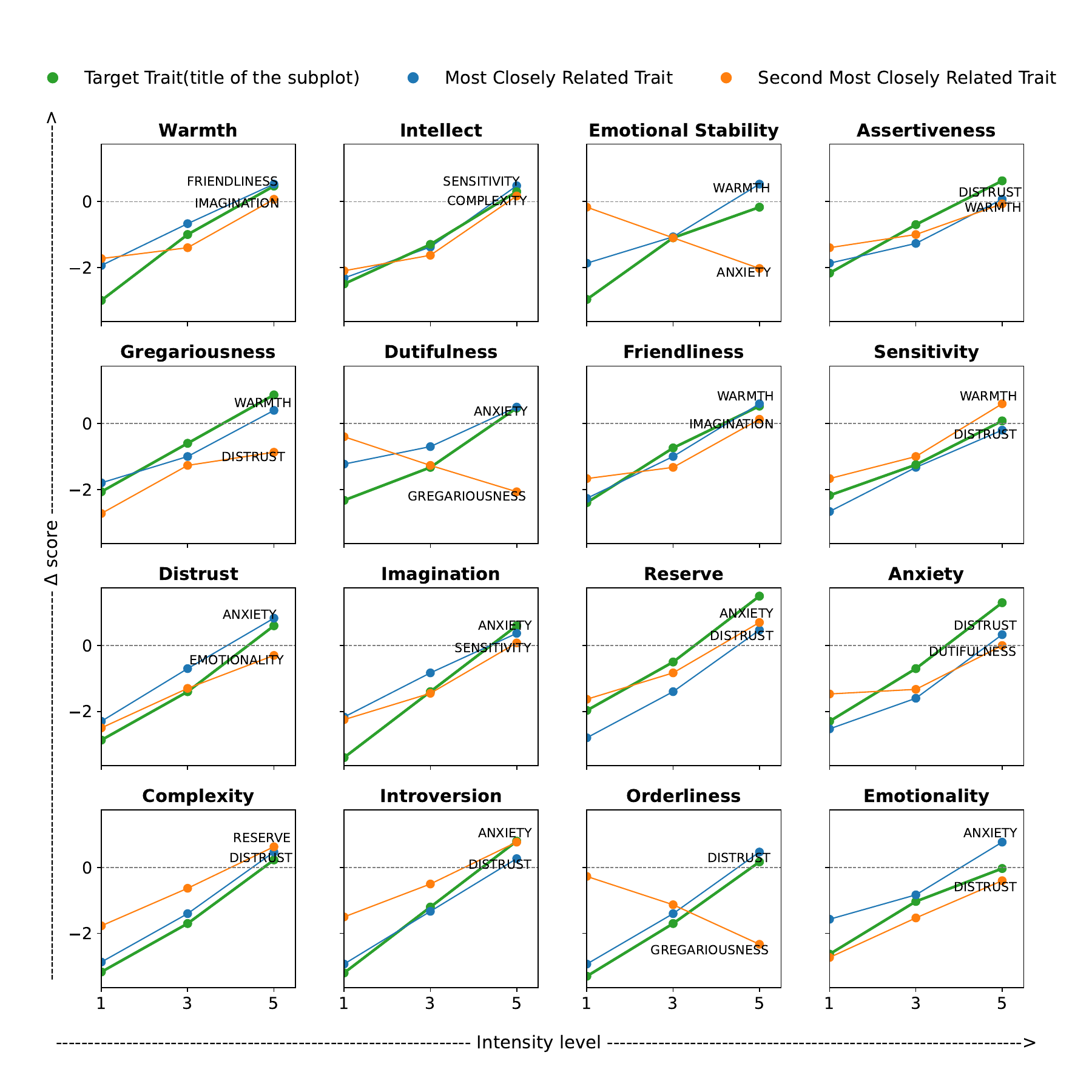}
    \vspace{0.5cm}
    \caption{\textbf{
    Target Trait Influence and Top Co-Movers in Mistral}. 
}
\label{fig:mistral-correlation-graph}
\vspace{0.5cm}
\end{figure*}

Figure \ref{fig:delta_all} illustrates the effect of inducing traits at intensity levels 1, 3, and 5, through deltas, allowing us to focus solely on the magnitude and direction of change. These bar charts provide a comprehensive comparison across all 16 traits for all four LLMs. As intensity increases, we observe a smooth and directional shift in trait deltas - from negative values at intensity 1, to moderate values closer to zero at intensity 3, and positive deltas at intensity 5. This trend indicates that SAC enables monotonic modulation of personality traits, with intensity 3 representing a plausible midpoint. This is a stark improvement from P2 Prompting wherein numerous negative deltas were observed even when positively keyed as seen in Tables \ref{tab:P2_deltas_part1} and \ref{tab:P2_deltas_part2}. The consistency of these patterns across traits and models reinforces the robustness and semantic alignment of our framework.
%CHECK TABLES AND FIGURE NAMES

Next, we analyse inter-trait movement across LLMs to understand whether or not LLMs are capable of emulating latent psychological structures known in human personality science.
\textbf{Co-movers} were selected by identifying the two traits with the largest deltas (positive or negative) accompanying the induction of a target trait as can be seen in Figure \ref{fig:gemini-claude-correlation-graph}. While Figures \ref{fig:gemini-claude-correlation-graph} shows the results for \texttt{Claude 3.7 Sonnet}, we observed similar modulation and co-movement patterns in the other three LLMs as well, indicating that the SAC framework generalizes well across architectures 
% jsc
(publicly accessible via Zenodo \cite{sac_zenodo}). 

The presence and behavior of these co-movers offer insights into whether LLMs capture inter-trait dependencies beyond the directly manipulated trait.

Table \ref{tab:co-mover} summarizes the top co-movers for each target trait based on the most reactive trait shifts observed across all four LLMs.
% The directionality of these relationships aligns with known psychological dynamics, providing further evidence of LLMs' latent trait modelling capabilities. 
Across all experiments, we observe that all four LLMs not only respond to individual traits with fidelity but also adjust semantically related traits in a manner consistent with human personality psychology. 
These structured co-movements observed in Table \ref{tab:co-mover} are not merely reactive but suggest that LLMs have internalized coherent personality architectures. 
Our intensity-control method reveals these underlying structures, and also enables precise and interpretable persona shaping.

\begin{table*}[t!]
\caption{Top co-movers and interpretive insights across traits.}
\label{tab:co-mover}
\centering
\scriptsize
\setlength{\tabcolsep}{4pt}
\renewcommand{\arraystretch}{1.15}
\resizebox{\textwidth}{!}{%
\begin{tabular}{|l|l|c|p{12cm}|}
  \hline
  \textbf{Target} & \textbf{Co-movers} & \textbf{Dir.} & \textbf{Interpretation} \\
  \hline
  Warmth & Reserve, Distrust & ↓ & Warmth promotes social openness and trust, thus dampening traits related to social withdrawal and skepticism. \\
  \hline
  Intellect & Sensitivity, Complexity & ↑ & Models associate cognitive richness with nuanced emotional and abstract thought. \\
  \hline
  Emotional Stability & Anxiety & ↓ & Affective regulation reduces psychological volatility. \\
  \hline
  Assertiveness & Reserve & ↓ & A rise in assertiveness reduces passivity and detachment. \\
  \hline
  Gregariousness & Introversion, Reserve & ↓ & Heightened sociability directly suppresses social withdrawal and reservation. \\
  \hline
  Dutifulness & Orderliness & ↑ & Reflects classical conscientiousness links between responsibility and structure. \\
  \hline
  Friendliness & Warmth, Gregariousness, Reserve & ± & Social engagement rises while social withdrawal falls. \\
  \hline
  Distrust & Anxiety, Reserve & ↑ & A guarded persona co-expresses with defensiveness and reduced openness. \\
  \hline
  Reserve & Warmth, Distrust & ± & Greater detachment increases skepticism and reduces interpersonal warmth. \\
  \hline
  Anxiety & Emotional Stability & ↓ & Internal reactivity and stability are opposing traits, consistently inverted. \\
  \hline
  Complexity & Intellect & ↑ & Trait pairs show abstract intellectual depth and complexity rising together. \\
  \hline
  Introversion & Reserve & ↑ & Solitude-seeking behaviors are intertwined in both direction and degree. \\
  \hline
  Emotionality & Anxiety, Reserve & ± & Emotional sensitivity partially aligns with internal turbulence. \\
  \hline
\end{tabular}
}
\end{table*}

\section{Limitations}
While our work advances personality modelling by introducing fine-grained intensity control through SAC prompting, several limitations remain. First, the reliance on self-reported questionnaire-style responses may not fully capture deeper behavioral tendencies in LLMs. Second, observed differences in trait expression across models may conflate genuine induction effects with inherent training data distributions or architectural biases. Third, while we focus on the 16PF framework, the generalizability of our method to alternative or hierarchical personality models remains an open question for future work. Addressing these challenges will help build more robust, adaptive, and psychologically grounded AI systems.

%%%%%%%%%%%%%%%%%%%%%%%%%%%%%%%%%%%%%%%%%%%%%%%%%%%%%%%%%%%%%%%%%%%%%%%%%%%
\section{Discussion and Summary of Observations}
% check p2 negative deltas line after sac confirm
This work introduces a shift from the widely used Big Five framework to Cattell’s more granular 16 Personality Factors (16PF) model, through the construction of a new benchmark: PERS-16. Compared to the Big Five, 16PF allows for more expressive and fine-grained personality modeling, enabling the evaluation and control of traits that are typically collapsed into broader categories - such as \textit{Sensitivity} and \textit{Orderliness} - in traditional models.
In our baseline evaluation using PERS-16, we observed that most traits naturally clustered around mid-scale values (around 3-3.5), suggesting neutral, balanced personalities. The differences across models were quantified using Euclidean distance, offering a psychometric lens for comparing LLM behavior.
Applying the P2 prompting method to PERS-16 revealed its limitations. While P2 could induce some traits, its binary nature and lack of intensity control resulted in inconsistent personality shifts, often producing flat or negative deltas even for positively keyed traits. This confirmed that P2 was insufficient for controlled personality modulation within a fine-grained trait space.
To overcome these issues, we proposed the Specific Attribute Control (SAC) framework, which enables trait induction across five intensity levels (1-5) along five interpretable behavioral dimensions. In SAC-neutral mode, models exhibited more stable and interpretable personality profiles than under P2. SAC-induced evaluations revealed smooth, graduated shifts in trait expression across all three LLMs. Notably, the emergence of consistent co-movers - traits that shifted in semantically or psychologically aligned ways with the target trait - highlighted the presence of latent, multidimensional personality structures within LLMs.

\section{Conclusion and Future Work}
Our study demonstrates that continuous, intensity-based personality control, enabled by the SAC framework and grounded in the 16PF model, offers a robust and interpretable method for modulating LLM personas. Through systematic evaluations across neutral baselines, binary induction (P2), and graded induction (SAC), we show that LLMs not only respond predictably to personality modulation but also reflect inter-trait dynamics aligned with human psychometric theory. This establishes a strong foundation for building adaptive, psychologically informed AI agents. 
Future work will explore multi-trait induction to simulate more human-like, multidimensional personalities. Another direction is real-time dynamic modulation of personality based on user interaction, enabling context-aware adaptations during dialogue.

\section*{\uppercase{Acknowledgements}}
% By using the \texttt{ack} environment to insert your (optional) 
% acknowledgements, you can ensure that the text is suppressed whenever 
% you use the \texttt{doubleblind} option. In the final version, 
% acknowledgements may be included on the extra page intended for references.
Portions of this manuscript were refined using AI-based writing assistants (e.g., ChatGPT) for grammar, clarity, and style. All scientific content, methodology, and analysis were developed and authored by the researchers.

%%%%%%%%%%%%%%%%%%%%%%%%%%%%%%%%%%%%%%%%%%%%%%%%%%%%%%%%%%%%%%%%%%%%%%%%
\bibliographystyle{apalike}
{\small
\bibliography{example}}

\end{document}